\documentclass[letterpaper]{article} 
\usepackage{aaai24}  
\usepackage{times}  
\usepackage{helvet}  
\usepackage{courier}  
\usepackage[hyphens]{url}  
\usepackage{graphicx} 
\usepackage{natbib}  
\usepackage{caption} 
\usepackage{colortbl}

\usepackage{graphicx}               
\usepackage{tabularx}               
\usepackage{booktabs}
\usepackage{amsmath}
\usepackage{stmaryrd}
\usepackage{xcolor}
\usepackage{soul}

\newcolumntype{C}{>{\centering\arraybackslash}X}
\usepackage{multirow}               
\usepackage{diagbox}                
\usepackage{hhline}                 
\usepackage{color}                  
\usepackage{amsmath}                
\usepackage{mathtools}              
\usepackage{enumitem}               

\usepackage{subfigure}
\usepackage{booktabs}
\usepackage{extarrows}
\usepackage{makecell}
\usepackage{xcolor}
\usepackage{algorithm}
\usepackage{algpseudocode}
\usepackage{latexsym}
\usepackage{inconsolata}
\usepackage{graphicx}
\usepackage{bbding}

\usepackage{soul}

\definecolor{RoseQuartzBg}{HTML}{F7CAC9}
\definecolor{RoseQuartz}{HTML}{F5A798}
\definecolor{Serenity}{HTML}{92A8D1}
\definecolor{OrangeRed}{rgb}{1.0, 0.27, 0.0}
\definecolor{Red}{rgb}{1.0, 0.0, 0.0}
\definecolor{Turquoise}{HTML}{0F4C81}
\usepackage{xparse}
\NewDocumentCommand{\lifu}{ mO{} }{\textcolor{OrangeRed}{\textsuperscript{\textit{Lifu}}\textsf{\textbf{\small[#1]}}}}
\NewDocumentCommand{\minqian}{ mO{} }{\textcolor{violet}{\textsuperscript{\textit{Minqian}}\textsf{\textbf{\small[#1]}}}}
\NewDocumentCommand{\zhiyang}{ mO{} }{\textcolor{Serenity}{\textsuperscript{\textit{Zhiyang}}\textsf{\textbf{\small[#1]}}}}
\NewDocumentCommand{\ying}{ mO{} }{\textcolor{teal}{\textsuperscript{\textit{Ying}}\textsf{\textbf{\small[#1]}}}}

\NewDocumentCommand{\jy}{ mO{} }{\textcolor{brown}{\textsuperscript{\textit{jy}}\textsf{\textbf{\small[#1]}}}}

\newcommand{\dataset}{\textsc{MultiScript}}

\usepackage{newfloat}
\usepackage{listings}
\DeclareCaptionStyle{ruled}{labelfont=normalfont,labelsep=colon,strut=off} 
\floatstyle{ruled}
\newfloat{listing}{tb}{lst}{}
\floatname{listing}{Listing}
\title{\textsc{MultiScript}: Multimodal Script Learning for Supporting Open Domain Everyday Tasks}
\author{
    Jingyuan Qi\equalcontrib,
    Minqian Liu\equalcontrib,
    Ying Shen,
    Zhiyang Xu,
    Lifu Huang    
}
\affiliations{
    Department of Computer Science, Virginia Tech\\

    jingyq1@vt.edu,
    minqianliu@vt.edu,
    yings@vt.edu,
    zhiyangx@vt.edu,
    lifuh@vt.edu
}

\usepackage{bibentry}

\begin{document}

\maketitle

\begin{abstract}
Automatically generating scripts (i.e. sequences of key steps described in text) from video demonstrations and reasoning about the subsequent steps are crucial to the modern AI virtual assistants to guide humans to complete everyday tasks, especially unfamiliar ones. However, current methods for generative script learning heavily rely on well-structured preceding steps described in text and/or images or are limited to a certain domain, resulting in a disparity with real-world user scenarios. To address these limitations, we present a new benchmark challenge -- \dataset{}, with two new tasks on task-oriented multimodal script learning: (1) multimodal script generation, and (2) subsequent step prediction. For both tasks, the input consists of a target task name and a video illustrating what has been done to complete the target task, and the expected output is (1) a sequence of structured step descriptions in text based on the demonstration video, and (2) a single text description for the subsequent step, respectively. Built from WikiHow, \dataset{} covers multimodal scripts in videos and text descriptions for over 6,655 human everyday tasks across 19 diverse domains. To establish baseline performance on \dataset{}, we propose two knowledge-guided multimodal generative frameworks that incorporate the task-related knowledge prompted from large language models such as Vicuna. Experimental results show that our proposed approaches significantly improve over the competitive baselines. 


\end{abstract}

\section{Introduction}


\begin{figure}[!t]
    \centering
    \includegraphics[width=0.43\textwidth]{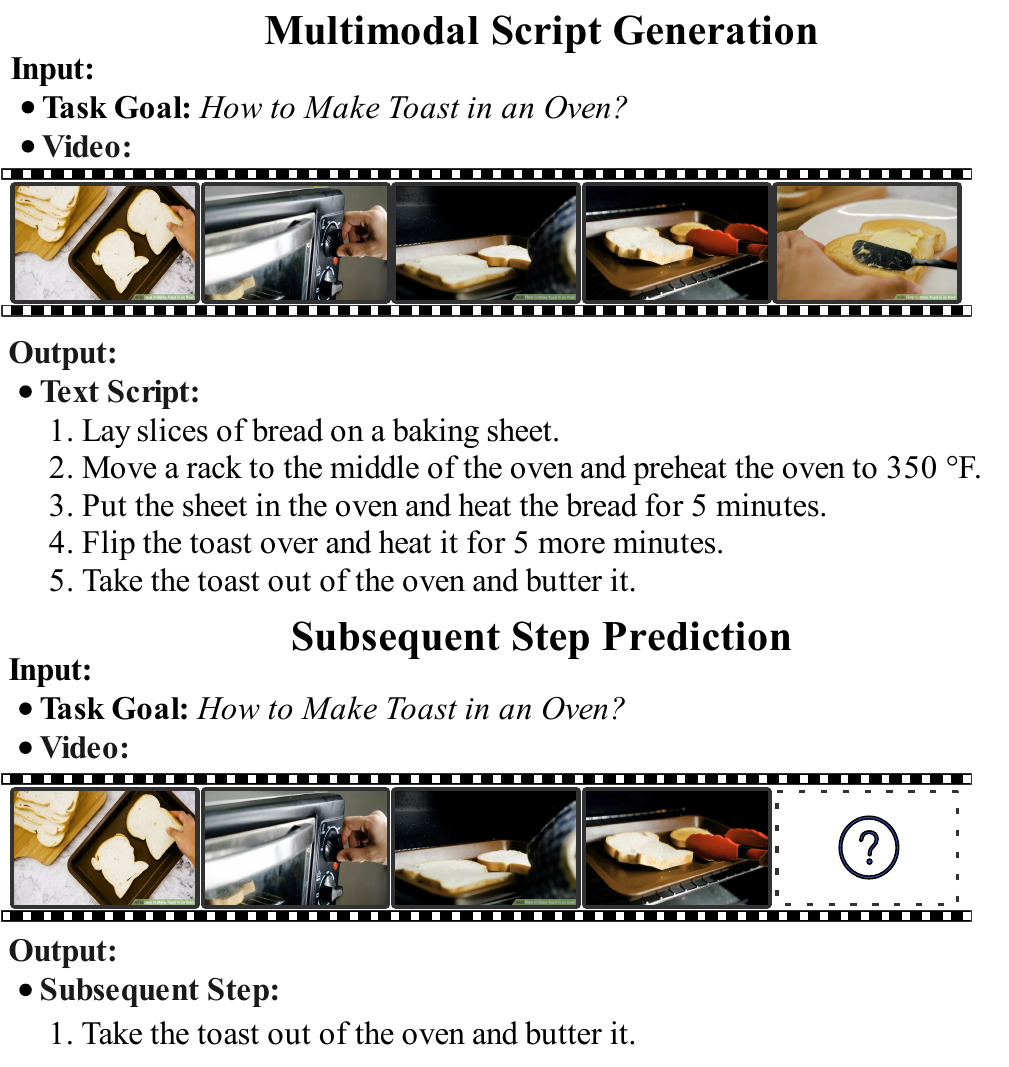}
    \caption{An example from \dataset{}. In order to generate the subsequent step, the model needs to understand the key steps from the video and gain sufficient instructional knowledge about ``how to make toast in an oven''.
    }
    \label{fig:intro_example}
\end{figure}

Recently, there has been an increasing focus on studies aimed at enabling AI assistants to better assist individuals in completing daily tasks, and two crucial aspects of these studies are to automatically: (1) provide a well-structured script, i.e., a sequence of standardized steps, as guidance for the target task; and (2) reason about the subsequent steps based on
what has been done by the individual.  Taking the task ``{\tt Make a toast in an oven}'' as an example, a script for achieving this task can be illustrated as: \textit{preheating the oven}, \textit{putting the sheet in the oven and heating the bread}, \textit{flipping the toast over}, and \textit{taking the toast out of the oven and buttering it}. While such script knowledge is typically authored by humans based on their experience or by watching video demonstrations, how to automatically acquire it and apply it to support the wide variety of everyday tasks is still an open research problem.



Existing studies on script learning can be categorized into two task formats: (1) \textit{multiple-choice question answering}, which is to select the most plausible subsequent step from a few candidates by giving a task goal and a preceding step~\cite{Yang2021}. Though such candidate-based formulation is a reasonable way to test models' capability of selecting the correct steps,
it is not feasible for real-world scenarios as the candidates are usually not available. (2) \textit{generative script learning}, which requires the model to generate a future step given a task goal and a sequence of preceding steps described in text and/or images~\cite{Lyu2020,Wang2022}. While this setup is more challenging, it assumes that the preceding steps are well-structured, which is not realistic either for actual AI assistants as in many use cases, they can only access preceding human actions from real-time or recorded video demonstrations that do not have clear boundaries between steps.  

To address these limitations and enable AI Assistants to better guide humans to complete everyday tasks, we propose a new benchmark challenge -- \dataset{} on task-oriented multimodal script learning with two novel tasks: (1) \textbf{Multimodal Script Generation}, which aims to extract and summarize all the key and necessary steps into structured text based on the target task and a video demonstration provided as input; and (2) \textbf{Subsequent Step Prediction}, of which the goal is to generate the most plausible and logical subsequent step based on the preceding steps demonstrated in an input video and the target task. \dataset{} consists of a large-scale dataset for each of the two tasks, covering a wide range of 6,655 everyday tasks across 19 domains, such as \textit{Finance}, \textit{Education}, etc. Figure~\ref{fig:intro_example} shows an example for each of the two tasks. We identify two critical challenges presented in \dataset{}: first, it requires accurate understanding and identifying all the key steps contained in the video as well as the target goal; second, it requires the model to gain sufficient background knowledge on various tasks and their necessary steps to correctly generate a complete script or a subsequent step for the target task, especially for the unseen tasks during inference.



To tackle these two challenges and establish baseline performance for the two tasks presented in \dataset{}, we propose a new knowledge-guided generative framework where we take a large-scale pre-trained language model (LLM) such as Vicuna~\cite{zheng2023judging}
as a task-agnostic universal knowledge source and dynamically induce and incorporate task-related knowledge to guide the model to generate a script or a subsequent step based on the input video demonstration. To relieve the effect of unrelated and/or incorrect external knowledge due to the randomness and hallucination of LLMs, we develop a natural language inference-based selector to selectively incorporate the prompt knowledge into the generation process.
Extensive experiments demonstrate that our proposed methods substantially outperform the competitive baselines\footnote{The codes, model checkpoints, and datasets are publicly available at \url{https://github.com/VT-NLP/MultiScript}.}. 

\section{Dataset Design}


\subsection{Task Formulation}

\paragraph{Multimodal Script Generation} 
To automate the acquisition of script knowledge from massive video demonstrations for human everyday tasks,
we introduce Multimodal Script Generation (MSG) task that aims to generate a structured text script by giving the task goal and a full video demonstration. More formally, we denote the task goal as $T$, input video demonstration as $V$, and the target text script as $\mathcal{S}$=\{$S_1$,...,$S_n$\} involving $n$ necessary and ordered steps. 
Compared to action anticipation \cite{Girdhar2021,zhong2022afft} or video captioning \cite{singh2020nits}, the generated scripts in our task are expected to be well-structured descriptions for a sequence of actions that follow a temporal and logical order.

\paragraph{Subsequent Step Prediction} 
To support the research of assisting humans to complete various daily tasks, we propose the Subsequent Step Prediction (SSP) task. 
Formally, given a task goal $T$ and a partial video $V_{i-1}$ that presents the preceding steps $\{S_1,...,S_{i-1}\}$ that have been completed, a model needs to predict a subsequent step $S_{i}$.
As there could be several subsequent steps following $S_{i-1}$, we define $S_{i}$ 
as the \textit{most plausible and logically reasonable step that is in a correct temporal order} given the task goal $T$ and preceding steps shown in $V_{i-1}$.


\subsection{Instructional Article Collection}

We use WikiHow\footnote{\url{https://www.wikihow.com/Main-Page}} as the source to build \dataset{} as it contains multimedia instructional articles, including key step descriptions in text and/or images and optional video demonstrations, for diverse open-domain human everyday tasks. To collect the WikiHow articles, we develop data collection programs based on \cite{zhang2020reasoning} and select 6,652 multimedia instructional articles, where each article must satisfy one of the following criteria: (1) the article contains a \textit{full video} $V^f$ demonstrating the whole process of completing the target task, or; (2) each step in the article is associated with an image or a video clip $C_i$ that demonstrates the process of completing a particular action. 
These 6,652 articles cover diverse human everyday tasks from 19 domains (e.g., \textit{food} and \textit{finance}). Appendix shows the detailed statistics for the 19 domains.

\subsection{Task Instance Construction}

We further process the multimedia articles and construct instances for each of the two tasks we proposed. 
Each article may involve one or multiple methods that share the same task goal but have different sequences of steps. When an article contains multiple methods and a full video $V^f$, $V^f$ likely covers all the methods. In this case, we further process $V^f$ to extract a video demonstration for each method. Specifically, we leverage the \textit{transition frames}\footnote{See Appendix for an example of transition frame.} 
in the video where each \textit{transition frame} indicates a transition occurs from one method to another.
Specifically, for each frame in $V^f$, we check if its pixel values in the content-fixed region
match those of the exemplar \textit{transition frame}. If so, we label it as a \textit{transition frame} and record its serial number as its position in $V^f$.
We then segment the video $V^f$  into $N$ segmented videos based on the transition frames. If $N$ is the same as the number of methods in the article, the segmented videos are sequentially associated with the methods. Otherwise, we deem the videos unavailable for these methods.
After the processing, for each method $M$, we extract the following information: (1) title of the article which is used as the target task goal $T$; (2) $n$ steps of text descriptions $\mathcal{S} = [S_1, ..., S_n]$ for the method $M$; (3) optional images or video clips for the $n$ steps $[C_1, ..., C_n]$; and (4) an optional video demonstration for the method $V^M$. 
We then construct the instances for the two tasks as follows.





\paragraph{Multimodal Script Generation}
For each method $M$, we create an instance that takes a task goal $T$ and a demonstration video $V$ as input, and a sequence of step descriptions $\mathcal{S}$ as the target output.
If each step $S_{i}\in\mathcal{S}$ is associated with a video clip, we concatenate the video clips for all the steps and use it as the demonstration video $V$ for $M$. If there is no video clip for any of the steps but a video demonstration $V^M$ is obtained from the article, we directly use $V^M$ as $V$. Otherwise, we will ignore this method. 
We finally obtained 6,169 instances for this task.


\paragraph{Subsequent Step Prediction} 
In this task, each instance contains a task goal $T$ and a partial video $V_{k}$ that demonstrates the $k$ proceeding steps as input, and a subsequent step description $S_{k+1}$ as target output. 
More specifically, for each method $M$ with $n$ steps of text descriptions $\mathcal{S} = [S_1, ..., S_n]$ and video clips $[C_1, ..., C_n]$, we take each step $S_{k+1}$  $(0\leq k\leq n-1)$ as the target output to create an instance where the input video $V_{k}$ is concatenated from $[C_1, ..., C_k]$. If any video clip in $[C_1, ..., C_k]$ is unavailable, we extract $V_{k}$ from the video demonstration $V^M$ for method $M$ by determining the end frame of step $S_k$ in the video. 
 The end frame is sourced from the last frame of video clip $C_k$ or the beginning frame of $C_{k+1}$. If neither of these two video clips is available, we will ignore this step. We iterate each frame in $V^M$ and seek matches for the end frames by comparing pixel values.
 If no match is found after the iteration, we cannot extract $V_k$ from $V^M$ and will ignore this step as a target step.
We finally collected 14,427 instances for this task. 

\paragraph{Train / Dev / Test Split} We split the instances created for each task into training, development, and test sets. 
For each task, to ensure the coverage of various domains in each set, we randomly sample 80\%, 5\%, and 15\% articles from each domain, and use the data instances created from them as the training, development, and test sets.
We name this benchmark challenge with two multimodal script learning tasks as \dataset{}. 
Table \ref{tab:generation_statistic} shows the statistics of the two tasks. 
\begin{table}[t]
\centering
\resizebox{0.4\textwidth}{!}{%
\small
\begin{tabular}{l|ccc}
\toprule \toprule
& Train          & Dev            & Test           \\
\midrule
\rowcolor[gray]{0.9}\multicolumn{4}{c}{Multimodal Script Generation} \\
\midrule 
\# Articles                         & 3,800           & 222            & 731            \\
\# Data Instance                        & 4,955          & 294            & 947           \\
Avg./Max. \# Step per Instance     & 7.3/54      & 7.7/43      & 7.5/32      \\
Avg./Max. Video Duration (s)        & 64/437       &69/459  &68/396  \\
\midrule 
\rowcolor[gray]{0.9}\multicolumn{4}{c}{Subsequent Step Prediction} \\
\midrule
\# Articles                         & 2,407           & 154            & 432            \\
\# Data Instance                        & 11,409          & 862           & 2,156           \\
Avg./Max. \# Preceding Steps     & 3.62 / 35      & 3.80 / 19      & 3.86 / 26      \\
Avg./Max. Video Duration (s)        & 33 / 977       & 34 / 439 & 36 / 568 \\\bottomrule
\bottomrule
\end{tabular}
}
\caption{Statistics of the two tasks in \dataset{}}
\label{tab:generation_statistic}
\end{table}

\section{Method}

\subsection{Overveiew}
\begin{figure*}[t]
    \centering
    \includegraphics[width=0.9\textwidth]{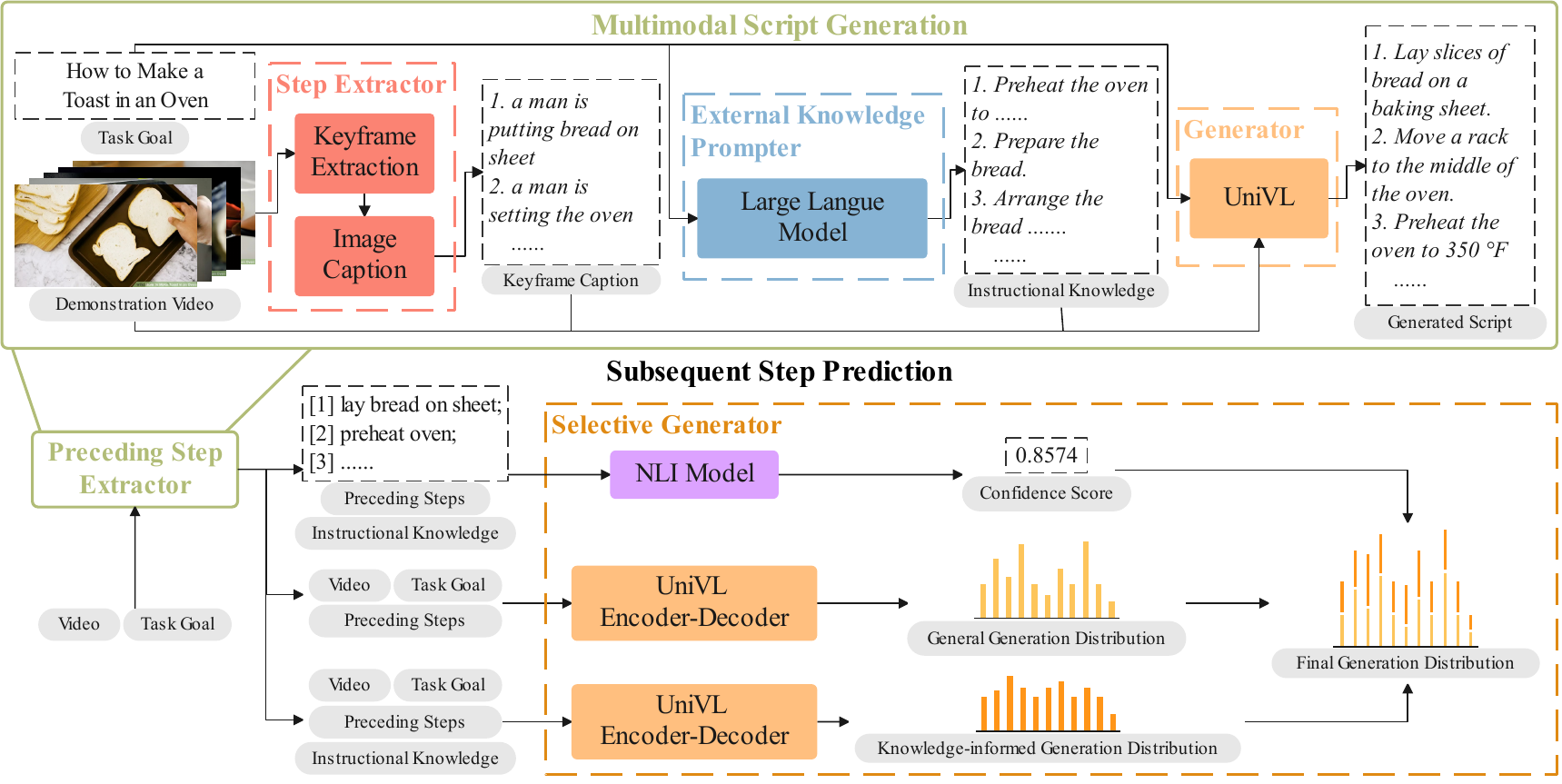}
    \caption{The overall framework to tackle our proposed two tasks. Three main components (1) a \textit{step extractor} to determine and extract steps demonstrated in video, (2) a \textit{external knowledge prompter} to acquire task-related knowledge, and (3) a \textit{generator}.
    } 

    \label{fig:method}
\end{figure*}

As shown in Figure \ref{fig:method}, we design two approaches for the two tasks introduced in \dataset{}. For multimodal script generation, 
we propose an approach consisting of three main components: (1) a \textit{Step Extractor} to extract a sequence of key frames from the input video and generate their captions to describe the sequence of key actions; (2) an \textit{External Knowledge Prompter} that induces task-specific instructional knowledge from large language models (LLMs) such as Vicuna; and (3) a \textit{Generator} to produce the final sequence of step descriptions by considering the input video, key frame captions as well as task-specific instructional knowledge. For subsequent step prediction, we first leverage the model traind on the previous multimodal script generation task to produce a script describing the preceding steps shown in the input partial video. Then, introduce a \textit{Selective Generator} to dynamically select external knowledge prompted from LLMs and incorporate them together with the preceding step descriptions to generate a subsequent step. In the following, we discuss the details of these two approaches.



\subsection{Multimodal Script Generation}

\subsubsection{Step Extractor}
Given a task goal $T$ and a demonstration video $V$, the step extractor aims to detect a sequence of key action frames from $V$ and generate their text descriptions. We employ Katna~\cite{katna}, an open-source tool, to extract a set of keyframes from $V$ and order them based on their timestamp in $V$, such that we obtain a sequence of chronologically ordered key frames $\mathcal{K}=\{K_1,\dots,K_m\}$, where $m$ is the total number of keyframes and is automatically determined by Katna~\cite{katna}. For each keyframe $K_i$, we further employ the pre-trained OFA model\footnote{We employ the checkpoint "OFA-Sys/ofa-base" in this work.}
\cite{wang2022unifying} to generate a caption, yielding a sequence of keyframe captions $\mathcal{I}=\{I_1,\dots,I_m\}$.


\subsubsection{External Knowledge Prompter}
LLMs have been shown to be able to capture generic instructional knowledge for open-domain human everyday tasks, which is valuable to inform the generator to produce a sequence of reasonable and logically coherent step descriptions from the input video. Considering this, we leverage an LLM, such as Vicuna~\cite{vicuna2023}, to produce task-specific instructional knowledge. 
Given the task goal $T$, we define a template prompt that instructs the LLM to generate a sequence of instructional steps \(\mathcal{P}=\{P_1,...,P_{n'}\}\), where $n'$ is the number of steps in the generated sequence.
The prompt includes an instruction of expert identity~\cite{xu2023expertprompting}, such as ``\textit{Imagine you are an expert on daily life tasks}'' at the beginning, followed by two in-context examples where each example is a pair of question and answer to demonstrate the expected output format. 
Following the prompt, we append a question based on the task goal $T$ to acquire a sequence of steps as the suggested procedures to complete the target task. 
Appendix
presents the template prompt we defined for the knowledge prompter. Note that the instructions prompted by the LLM are typically not precisely aligned with the input demonstration video, but they can provide a general workflow to complete the target task.

\subsubsection{Generator}

Given the input task goal $T$, video keyframe captions $\mathcal{I}=\{I_1,\dots,I_m\}$, instructional knowledge from LLMs $\mathcal{P}=\{P_1,...,P_{n'}\}$, as well as the demonstration video $V$, the generator aims to produce a script, $\mathcal{S}=\{S_1,...,S_n\}$, consisting of a sequence of action descriptions that can instruct humans to successfully complete the target task. Here, we employ UniVL~\cite{UniVL}, a pre-trained multimodal encoder-decoder model, as the generator. Specifically, we first concatenate all the text inputs $T, \mathcal{I}, \mathcal{P}$ with a split token {\tt [SEP]} and encode the concatenated sequence using the text encoder of UniVL. The video $V$ is first encoded as spatial and temporal representation feature~\cite{xie2018rethinking} by a separable 3D CNN (S3D) model~\cite{miech19howto100m}. The S3D features for the video $V$ are then encoded by the video encoder of UniVL. A cross encoder further takes in the features from the text encoder and video encoder and captures the interactions between the two modalities. Finally, we employ a Transformer-based decoder to produce a sequence of step descriptions $\mathcal{S}$.


\subsection{Subsequent Step Prediction}
\subsubsection{Preceding Step Extractor}

Unlike prior studies on task planning~\cite{Wang2022} which heavily rely on well-defined preceding steps, one significant challenge of our subsequent step prediction task lies in automatically inferring the preceding steps solely based on the input video. This objective is similar to the multimodal script generation task that produces a sequence of step descriptions from the input video. Considering this, we directly employ a model trained on the multimodal script generation task to generate a sequence of preceding steps $\mathcal{S}_k=[S_1, ..., S_k]$ given the task goal $T$ and a partial video $V_k$. 



\subsubsection{Adaptive Knowledge-informed Generator}
We further generate a subsequent step by considering the input task goal $T$, preceding steps $\mathcal{S}_k=[S_1, ..., S_k]$, the partial video $V_k$, as well as the instructional knowledge $\mathcal{P}=\{P_1,...,P_{n'}\}$ prompted from the LLM during the multimodal script generation. The prompted knowledge can provide the instructions related to the generator. However, it may not be well aligned with the preceding steps demonstrated in the input video since the prompted knowledge from LLMs can be too generic.
To adaptively incorporate the prompted knowledge, we propose to generate the subsequent step by producing two separate probability distributions at each decoding step, where one distribution considers the instruction knowledge as input and the other does not. 
Then, we fuse these two distributions with a dynamic confidence score.
Specifically, we employ two separate multimodal encoders based on UniVL: one encodes the concatenation of task goal $T$ and preceding steps $\mathcal{S}_k=[S_1, ..., S_k]$ with a split token {\tt [SEP]} as well as the spatial and temporal representation of the partial video $V_k$, and the other further considers the instructional knowledge $\mathcal{P}=\{P_1,...,P_{n'}\}$ as part of the textual input. 
At each decoding step, based on the encoded features of the two separate UniVL encoders, we produce two separate probability distributions: $D_g$ and $D_k$. 

To combine the two probability distributions at each decoding step, we need to characterize how likely the prompted knowledge is aligned and beneficial to the subsequent step generation. Here, we design an approach based on natural language inference (NLI) to first predict a confidence score for each individual step included in the prompted knowledge and then output an overall score as the weight to combine the two distributions. The confidence score for each step indicates how likely the step follows the preceding steps demonstrated in the input video. Specifically, we take the concatenation of the task goal $T$ and preceding steps $\mathcal{S}_k$ as a premise, each step in the instructional knowledge $P_i$ as a hypothesis, and use a pre-trained Deberta~\cite{he2020deberta}
\footnote{In this work, we employ the Deberta model with checkpoint "nli-deberta-v3-base".} as the backbone NLI model to predict a probability for the \textit{Entailment} label, which is then used as the confidence score. 
Note that to fine-tune the NLI model on our own dataset, we use the gold subsequence step as the positive hypothesis and sample a set of negative hypotheses from 
both preceding and future steps. Here, the ``future steps'' signify the steps that follow, but aren't immediately after, the preceding step.
Finally, we use the highest confidence score $c$ among all the steps as the weight to combine the two distributions: 
\begin{equation*}
    D_f = D_g + c*D_k
\end{equation*}
Based on the combined distributions $D_f$ at each decoding step, a Transformer-based decoder further predicts the output token and finally generates a subsequent step $S_{k+1}$.

\section{Experiment Setup}
\label{sec:experiment}




\paragraph{Evaluation Metrics}
We evaluate our methods and baselines with eight standard metrics: BLEU~\cite{Papineni2002} including BLEU-1, BLEU-2, BLEU-3 and BLEU-4, METEOR~\cite{Denkowski2014}, ROUGE-L~\cite{Lin2004}, BERTScore~\cite{Zhang2019}, and Semantic Textual Similarity (STS)~\cite{thakuretal2021}. Considering the variability in describing steps or task scripts, semantic similarity-based metrics hold greater significance in the proposed tasks.


\paragraph{Baselines}

For multimodal script generation, we first employ the video keyframe captions as the output script, serving as the most naive baseline (\textbf{Keyfram Caption}). Additionally, we assess the generic instructions prompted from the large language model, Vicuna-13B-1.1~\cite{vicuna2023}, as another baseline, to gauge the quality of the instructional knowledge (\textbf{Vicuna}). 
For subsequent step prediction, we compare our approach with two text-only based generation models, T5-Large with 7.7M parameters (\textbf{T5}) and Vicuna-13B-1.1~\cite{vicuna2023} (\textbf{Vicuna}), which take in the task goal $T$ and the preceding steps $\mathcal{S}_{k}$ and generates a subsequent step $S_{k+1}$. Both T5 and Vicuna are fine-tuned on the same dataset we created for subsequent step prediction.
We also compare our approach with several ablated variants, including:
\textbf{(1) UniVL}: directly takes in the task goal and input video, and generates a script or a subsequent step; \textbf{(2) UniVL+Step}: Besides the task goal $T$ and input video, it also takes in the steps demonstrated in the video. Specifically, for multimodal script generation, we use the video keyframe captions $\mathcal{I}$ as the step description, while for subsequent step prediction, we use the preceding steps $\mathcal{S}_k$ generated by a well-trained model for multimodal script generation; \textbf{(3) UniVL+Knowledge}: Besides the task goal $T$ and input video, it also considers the instructional knowledge prompted from large language models for both tasks; \textbf{(4) UniVL+Step+Knowledge}: complete variants of our proposed approaches for both tasks that consider the task goal $T$, input video, step descriptions generated from the input video as well as the instructional knowledge from the large language models.

\paragraph{Human Evaluation} In order to comprehensively understand the challenges of the proposed tasks and assess the disparity between machine and human performance, we also conduct a human evaluation for both tasks. We hired five graduate students with decent NLP backgrounds as annotators to independently complete the multimodal script generation and step prediction tasks on 100 randomly sampled instances. For the multimodal script generation task, annotators are provided the task goal $T$ and a demonstration video $V$, and were asked to summarize the script from the video. For the step prediction task, annotators were given the task goal $T$ and a partial demonstration video $V_k$, and were asked to predict the immediate subsequent step.

\section{Results and Discussions}
\subsection{Quantitative Comparison}
Table~\ref{tab:baseline_result_both} show the performance of both baselines and our approaches for two proposed tasks. In most cases, our proposed approaches outperform all baselines and their variants with significant margins.

\begin{table*}[]
\centering
\small
\resizebox{0.8\textwidth}{!}{%
\small
\begin{tabular}{l|ccccccccc}
\toprule\toprule
\textbf{Model}        & \textbf{BLEU-1}  & \textbf{BLEU-2}  & \textbf{BLEU-3}  & \textbf{BLEU-4}  & \textbf{METEOR} & \textbf{ROUGE-L} & \textbf{BERTScore} & \textbf{SBert} & \textbf{Average} \\ \midrule
\rowcolor[gray]{0.9}\multicolumn{10}{c}{Multimodal Script Generation} \\ \midrule
Keyframe Caption       & 21.03          & 4.93          &  0.95         & 0.17          & 15.58            & 19.82           & 83.84              & 45.05              & 23.92              \\
Vicuna        & 28.38          & 20.63          &  14.74         & 5.78          & 28.97            & 24.56           & 86.76              & 60.61              & 33.80                  \\\midrule
UniVL              & 34.60          & 26.50          &  20.11         & 10.39          & 31.75            & 35.30           & 88.32              & 63.39              & 38.8          \\
UniVL+Step             & \underline{42.74}          & \underline{32.61}          & \underline{24.57}          & \underline{11.71}          & \underline{38.01}            & \textbf{38.39}           & \underline{88.44}              & 61.45              & \underline{42.24}          \\
UniVL+Knowledge          & 36.33          & 27.68          & 20.93          & 11.00          & 33.32            & 34.95           & 88.32              & \underline{64.48}              & 39.63          \\
UniVL+Step+Knowledge           & \textbf{44.35}          & \textbf{33.84}          & \textbf{25.63}          & \textbf{13.24}          & \textbf{38.54}            & \underline{38.04}           & \textbf{88.98}              & \textbf{65.14}              & \textbf{43.47}           \\\midrule
Human Performance           & 22.20          & 7.86          & 2.94          & 0.51          & 20.92            & 28.21           & 88.33              & 67.89              & 29.78           \\ \midrule
\rowcolor[gray]{0.9}\multicolumn{10}{c}{Subsequent Step Prediction} \\ \midrule
T5       & 7.63          & 0.38          &  0.00        &  0.00         &  5.42           & 11.81           & 80.85              &  14.74             & 15.10             \\
Vicuna        & 9.32          &  2.94         &  0.38         & 0.1          & 12.06            & 11.94           & 83.49              &  34.78             & 19.38                   \\\midrule
UniVL              & 36.46          & 25.45          &  4.44         & 2.36          & 39.96            & 40.19           & 89.50              & 51.17              & 36.19          \\
UniVL+Step             & 36.64          & 25.80          & 4.59          & 2.52          & 39.90            & 40.03           & 89.54              & 51.80              &  36.35         \\
UniVL+Knowledge          & \underline{37.08}        & \underline{26.04}          & \underline{4.63}          & \underline{2.52}         & \underline{39.98}            & \textbf{40.86}           & 89.7              & \underline{52.04}              &   \underline{36.60}        \\
UniVL+Step+Knowledge           & \textbf{37.57}          & \textbf{26.45}          & \textbf{5.45}          & \textbf{3.00}          & \textbf{40.61}            & \underline{40.73}           & \textbf{89.7}             & \textbf{53.59}              & \textbf{37.14}           \\\midrule
Human Performance           & 8.63          & 2.00          & 0.83          & 0.00          & 7.55            & 11.77           & 85.76              & 37.71              &  19.28          \\
\bottomrule\bottomrule
\end{tabular}
}
\caption{Automatic evaluation results on multimodal script generation and subsequent step prediction tasks. All metrics are reported in percentage (\%). We highlight the best scores in bold and the second best with underline. }
\label{tab:baseline_result_both}
\end{table*}

\subsubsection{Multimodal Script Generation}
From the results of the \textbf{Keyframe Caption} baseline in Table \ref{tab:baseline_result_both}, it is evident that for multimodal script generation, relying solely on video input and merely captioning the keyframes of the demonstration video does not accurately summarize the video content. Fine-grained information is missing when extracting the keyframes, and details of the keyframe images are compromised when converting them into text captions. In addition, scripts generated by LLMs based solely on task goals also lack accuracy, as shown by the performance of \textbf{Vicuna}.
Due to the diverse solutions available for a human everyday task, solely relying on the task goal, Vicuna can only produce generic scripts that may not align with the methods demonstrated in the video, leading to scripts deviating significantly from the ground truth.

The results from the middle group of Table \ref{tab:baseline_result_both} present the performance of the variants of our approach. Our models that incorporate external knowledge (\textbf{UniVL+Knowledge} and \textbf{UniVL+Script+Knowledge}) provide average absolute gains of 0.83\% and 4.67\%, respectively, over the standard \textbf{UniVL} model, which demonstrates the benefit of the task-specific instructional knowledge prompted from LLMs. An interesting finding is that, the model that integrates only keyframe captions as the supplementary input (\textbf{UniVL+Step}) also exhibits remarkable improvements, surpassing even the knowledge-enhanced variant \textbf{UniVL+Knowledge}. The possible reasons are (1) the key frames extracted from the video usually highlight the crucial steps to complete the target task, and (2) the instructions prompted by LLMs are not aligned with the steps demonstrated in the input video. It might elucidate why the improvements benefiting from the step extraction component do not mirror its significant performance in the multimodal script generation task.

\subsubsection{Subsequent Step Prediction}
From Table \ref{tab:baseline_result_both}, it is evident that the approaches based on single modality(\textbf{T5} and \textbf{Vicuna}) underperform all multimodal variants of our approach for subsequent step prediction, e.g., the baseline \textbf{UniVL} outperforms \textbf{T5} and \textbf{Vicuna} by 21.9\% and 16.81\%, respectively. A potential reason is that the preceding steps produced by the multimodal script generation approach are not fully correct so the errors are accumulated in the subsequent step generation, while the multimodal variants of our approach can alleviate or correct the errors in preceding steps based on the input video. Among the variants of our approach, \textbf{UniVL+Knowledge} outperforms both \textbf{UniVL} and \textbf{UniVL+Step} because (1) \textbf{Vicuna}, being substantially larger than \textbf{UniVL} and trained on a more extensive dataset, serves as a more robust knowledge source, enabling the generation of more precise and fine-grained steps. Although the instructional knowledge does not fully match the demonstrated method, it may share similar steps. Combined with the selective mechanism, this overlap allows the generator to integrate meaningful information while reducing the impact of misaligned instructional knowledge; (2) the extraction of the preceding steps, as mentioned earlier, is not fully correct, leading to misrepresentations of the task state.

\begin{figure}[t]
    \centering
    \includegraphics[width=0.48\textwidth]{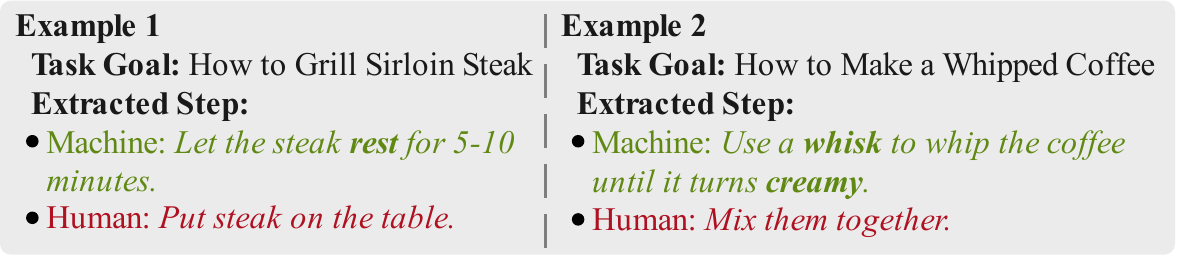}
    \caption{Example of extracted steps by human and machine in multimodal script generation task.}
    \label{fig:professional}
\end{figure}

\paragraph{Human Evaluationn}
Table~\ref{tab:baseline_result_both} reveals that, for multimodal script generation, machines outperform humans in terms of word accuracy and professionalism.
The machine employs precise technical terminology for action descriptions and offers more standardized and accurate portrayals of entities and their states, e.g., \textit{"rest", "whisk", "creamy"} in Figure \ref{fig:professional}, which is a challenge for humans without expertise.
However, humans significantly outperform machines based on semantic-based evaluation. One possible reason is that humans might not recognize specific tool or entity names from a demonstration video, but it is easy for them to clearly elucidate the workflow to complete the target task. 
For subsequent step prediction tasks, humans consistently underperform machines across all metrics. This can be attributed to the essentiality of domain-specific knowledge in this task. In multimodal script generation, humans can describe the content demonstrated in the video and effectively achieve high semantic alignment with the golden script. However, in the subsequent step prediction task, human annotators without domain knowledge struggle to grasp the workflow and steps involved in an everyday task. Consequently, their predictions frequently diverge from the correct subsequent steps.

\begin{figure*}[t]
    \centering
    \includegraphics[width=0.9\textwidth]{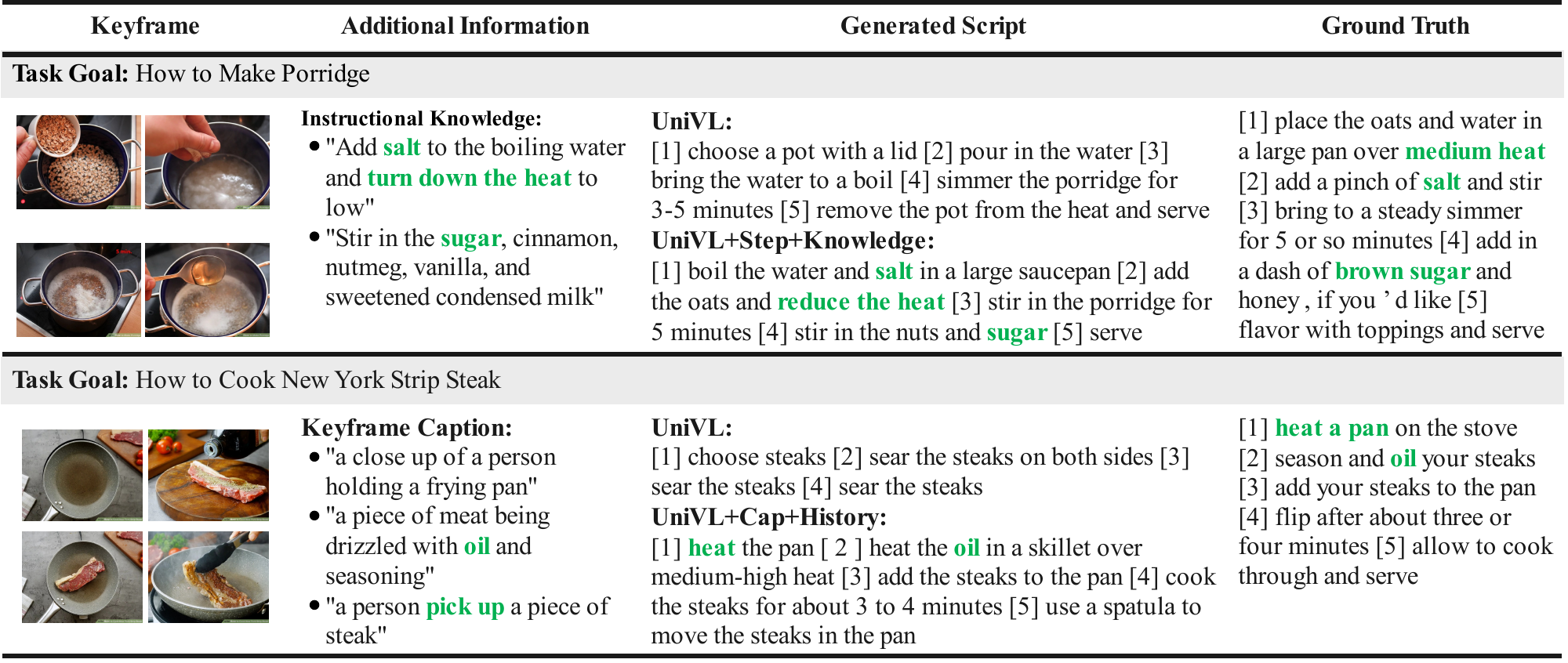}
    \caption{Qualitative result of different model variants in multimodal script generation task. The green highlights indicate key factors for better generations.} 
    \label{fig:result_exp}
\end{figure*}
\begin{figure*}[t]
    \centering
    \includegraphics[width=0.9\textwidth]{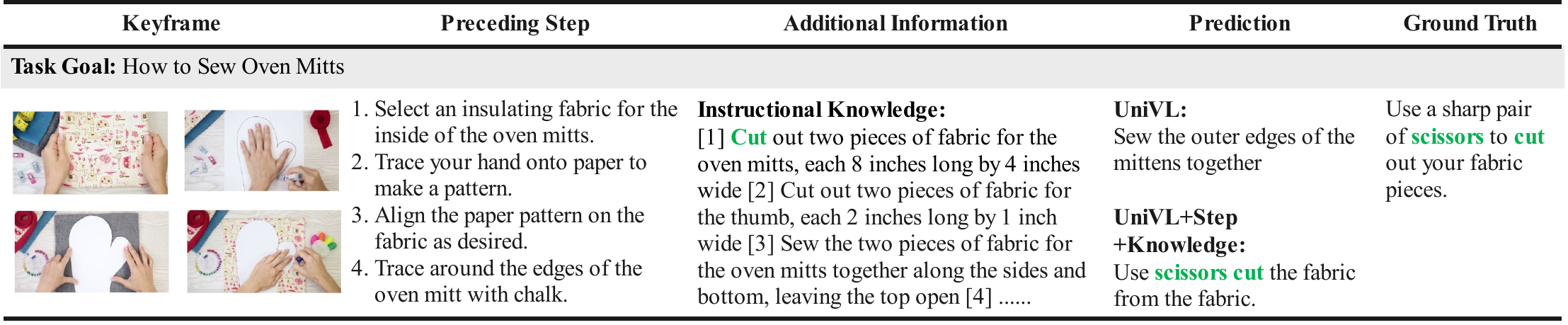}
    \caption{Qualitative result of different model variants in subsequent step prediction task. The green highlights indicate key factors for better generations.} 
    \label{fig:result_exp2}
\end{figure*}
\subsection{Qualitative Analysis}
Figure~\ref{fig:result_exp} and Figure~\ref{fig:result_exp2} present the qualitative results from various model variants in multimodal script generation and subsequent step prediction tasks, showing approaches that incorporate external instructional knowledge and video keyframe captions substantially enhance the quality of generation and prediction. 
Additional instructional knowledge provides finer-grained commonsense details that are hard to discern from demonstration videos. For instance, entities with comparable appearances, e.g., “\textit{salt}”, “\textit{sugar}” in Figure~\ref{fig:result_exp}, are not easily distinguished from the video. With clear clarification and definitions from the instructional knowledge, our approach is able to extract more accurate and fine-grained steps. 
For the subsequent step prediction task, a well-defined workflow outline aids the model in discerning the current task stage. As demonstrated in Figure~\ref{fig:result_exp2}, our approach accurately predicts the“\textit{cut}” step, reflecting a deeper understanding of the task flow compared with UniVL.

\section{Related Work}

\paragraph{Action Anticipation}
In computer vision, action anticipation~\cite{Girdhar2021,zhong2022afft,Roy2022} task aims to predict future actions based on past frames. A challenge of this task is discerning human patterns from these actions. 
In contrast, task-oriented generation has fixed steps that need to be followed. Given the pre-define task goal, the model should produce consistent outputs even with noisy historical actions.

\paragraph{Multimodal Script Learning}
Several multimodal approaches~\cite{Narasimhan2022,Wang2022,multiinstruct,qi-etal-2023-art} have been proposed in recent years to tackle script learning tasks. While there is significant progress in predictive performance, several practical issues have yet to be adequately addressed. For example, in candidate-based tasks such as classification~\cite{Lin2022} and multiple-choice~\cite{Yang2021}, the acquisition of reliable candidates is often unstable due to the lack of user input and the limited number of documented tasks. 


\paragraph{Datasets Related to Future Prediction Task}
Three types of datasets have been used in previous research on future prediction tasks: visual-only~\cite{Damen2022RESCALING,Damen2018EPICKITCHENS,Li2018}, text-only~\cite{puig2018virtualhome,lyu2021goal,le-etal-2023-improved}, and multimedia datasets~\cite{yang2021visual,tang2019coin,miech2019howto100m,visionFlan2023}. Our dataset distinguishes itself from these previous works in two aspects: (1) Compared with visual/text-only dataset, our dataset is a multimedia dataset, comprising video, image, and text descriptions for each instructional step. The data inclusion from multiple modalities provides unique and complementary information
\footnote{See Appendix for data type comparison of existing work and \dataset{}.}.
(2) Text description of our dataset is more fine-grained, using original instructional article step descriptions from WikiHow rather than using the transcript or summary annotations of video. 

\section{Conclusion}
In this paper, we introduced a new benchmark challenge encompassing two task-oriented multimodal script learning tasks: multimodal script generation and subsequent step prediction. To support further research in this domain, we present a new dataset consisting of over 6,655 everyday human tasks across 19 domains. We proposed a knowledge-informed framework that adaptively integrates task-related instructional knowledge into the multimodal generative model. While our approach demonstrated encouraging results, there are still limitations and challenges that need to be addressed in future work on these tasks.

\section*{Acknowledgments}
This research is based upon work supported by the U.S. DARPA ECOLE Program \# HR001122S0052. The views and conclusions contained herein are those of the authors and should not be interpreted as necessarily representing the official policies, either expressed or implied, of DARPA or the U.S. Government. The U.S. Government is authorized to reproduce and distribute reprints for governmental purposes notwithstanding any copyright annotation therein.

\bibliography{aaai24}


\appendix
\newpage
\section{Appendices}

\subsection{Distribution of Multimedia Instructional Articles in Different Domains}
\label{apx:domain_detail}
\begin{figure}[h]
    \centering
    \includegraphics[width=0.4\textwidth]{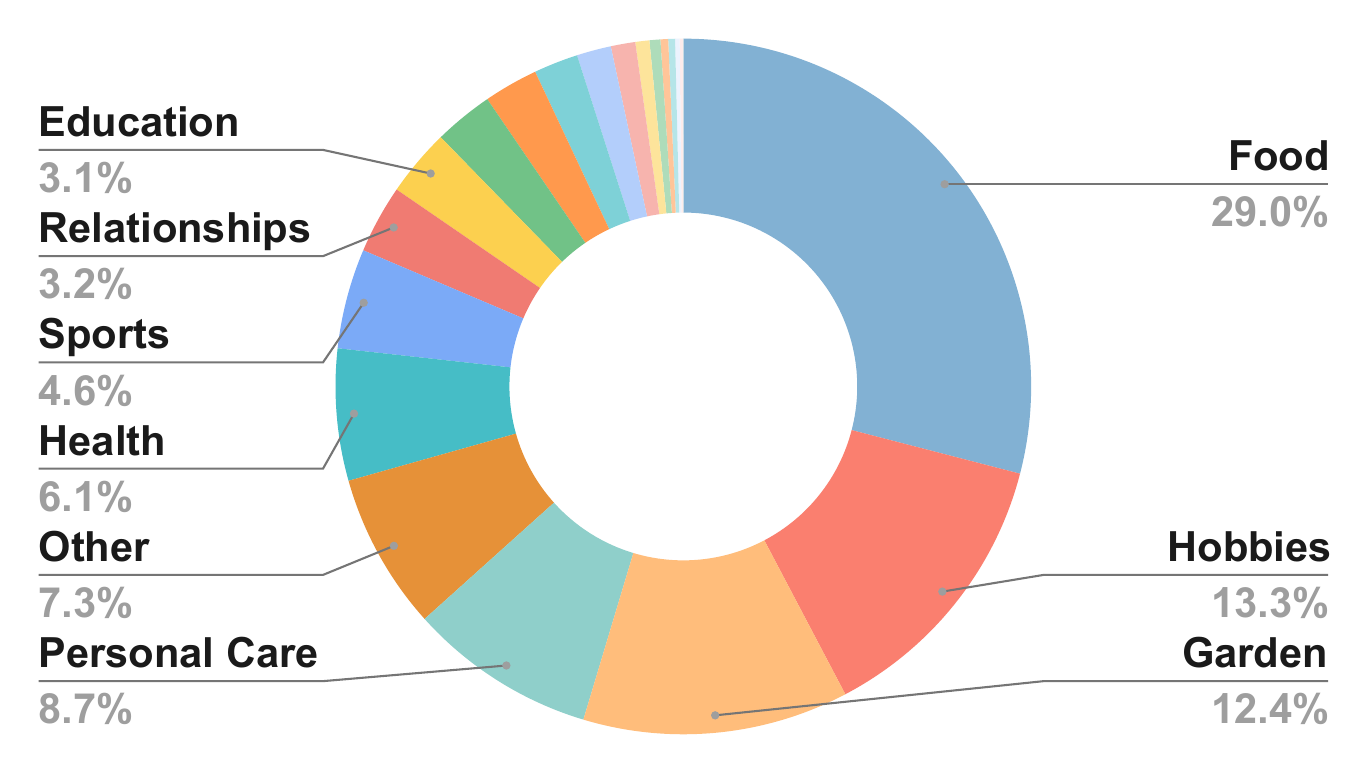}
    \caption{The distribution of multimedia articles in different domains in \dataset{} for multimodal script generation task.}
    \label{fig:article_domian_distribution_summary}
\end{figure}

\begin{figure}[h]
    \centering
    \includegraphics[width=0.4\textwidth]{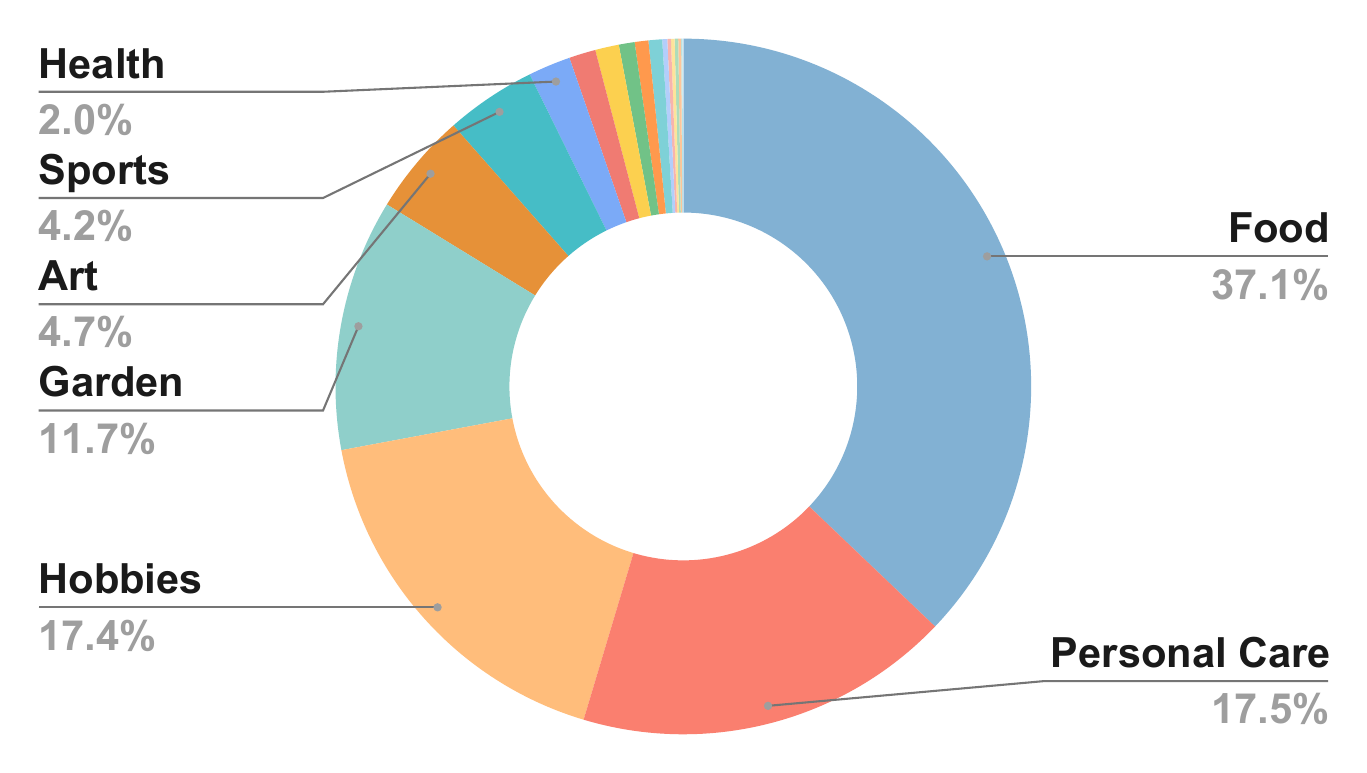}
    \caption{The distribution of  multimedia instructional articles in different domains in \dataset{} for subsequent step prediction task.
    }
    \label{fig:article_domian_distribution_prediction}
\end{figure}

In \dataset{}, we collected instructional articles on human everyday tasks spanning over 19 domains. As one can observe from Figure~\ref{fig:article_domian_distribution_summary},~\ref{fig:article_domian_distribution_prediction}, the \textit{Food and Entertainment} domain constitutes over a quarter of all articles. Besides that, in the subsequent step prediction task, both the \textit{Personal Care and Style} and \textit{Home and Garden} domains account for roughly 20\% of the articles. In contrast, there is a more balanced article domains distribution for the multimodal script generation task: \textit{Hobbies and Crafts}, \textit{Home and Garden}, and \textit{Personal Care and Style} each make up about 10\% of the articles.

\begin{table}[t]
\resizebox{0.5\textwidth}{!}{%
\begin{tabular}{c|cc}
\toprule\toprule
\textbf{Domain}             & \textbf{\begin{tabular}[c]{@{}c@{}}Multimodal Script\\ Generation\end{tabular}} & \textbf{\begin{tabular}[c]{@{}c@{}}Subsequent Step\\ Prediction\end{tabular}} \\\midrule
Animals and Pets            & 120                                                                             & 22                                                                            \\
Art and Entertainment       & 131                                                                             & 142                                                                           \\
Business and Finance        & 31                                                                              & 5                                                                             \\
Cars \& Other Vehicles      & 76                                                                              & 37                                                                            \\
Computers and Electronics   & 98                                                                              & 19                                                                            \\
Education and Communication & 149                                                                             & 33                                                                            \\
Family Life                 & 17                                                                              & 4                                                                             \\
Food and Entertaining       & 1380                                                                            & 1111                                                                          \\
Health                      & 291                                                                             & 59                                                                            \\
Hobbies and Crafts          & 630                                                                             & 522                                                                           \\
Holidays \& Traditions      & 24                                                                              & 19                                                                            \\
Home and Garden             & 587                                                                             & 350                                                                           \\
Insect pests and diseases   & 7                                                                               & 0                                                                             \\
Other                       & 346                                                                             & 0                                                                             \\
Personal Care and Style     & 414                                                                             & 524                                                                           \\
Philosophy and Religion,    & 15                                                                              & 7                                                                             \\
Relationships               & 151                                                                             & 5                                                                             \\
Sports and Fitness          & 221                                                                             & 126                                                                           \\
Travel                      & 1                                                                               & 2                                                                             \\
Work World                  & 10                                                                              & 1                                                                             \\
Youth                       & 54                                                                              & 5\\\bottomrule\bottomrule
\end{tabular}
}
\caption{Detail of distribution of multimedia instructional articles in different domains} 
\label{tab:domain_detail}
\end{table}


\subsection{Transition Frame}
\begin{figure}[H]
    \centering
    \includegraphics[width=0.35\textwidth]{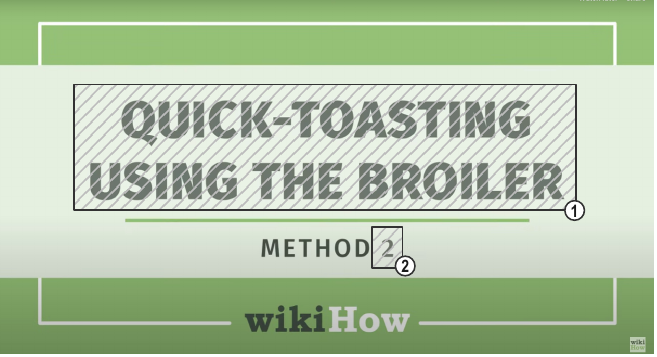}
    \caption{Example of the transition frame, which signals a shift from one method to another. Apart from \textcircled{\small{1}} method name and \textcircled{\small{2}} method ID, all other content remains constant with fixed pixel values.}
    \label{fig:transition_frame}
\end{figure}

We employ the \textit{transition frame} to identify method transitions in the demonstration video. The content within the unshaded region of Figure~\ref{fig:transition_frame} remains constant for all \textit{transition frames}. To detect \textit{transition frames}, we iterate each frame in the video \(V^f\) and compare the pixel values of its content-fixed region with those of the \textit{transition frame}'s content-fixed region.

\subsection{Template Prompt Defined for Knowledge Prompter}
\begin{figure}[h]
    \centering
    \includegraphics[width=0.45\textwidth]{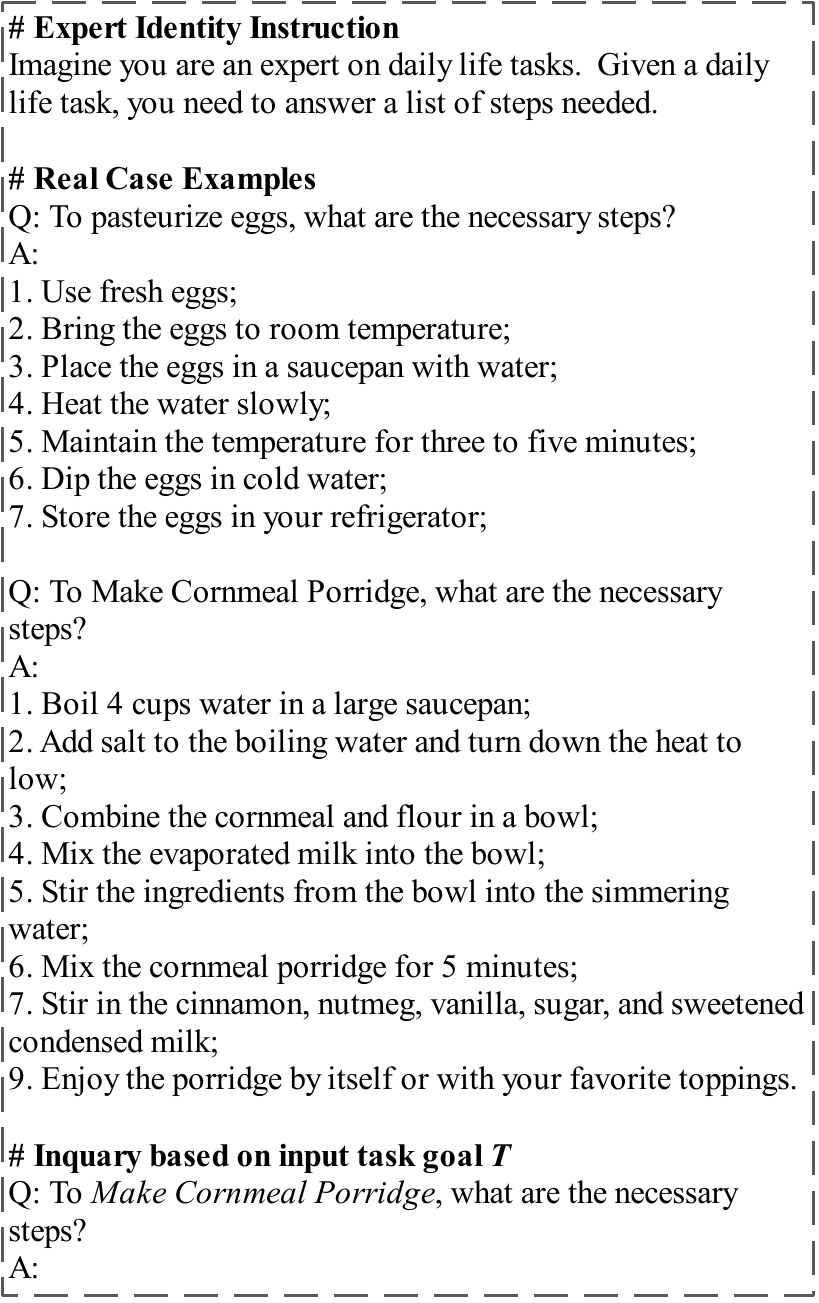}
    \caption{Template prompt used to acquire task-related instructional knowledge.}
    \label{fig:prompt}
\end{figure}

We define a template to elicit external instructional knowledge from LLMs. This template begins with an expert identity instruction, such as ``\textit{Imagine you are an expert on daily life tasks},'' followed by two real case examples. Each example presents a pair of question and answer to demonstrate the expected output format. Figure~\ref{fig:prompt} provides a detail of the template prompt.

\subsection{Comparison of Datasets}
Figure~\ref{tab:exist_dataset} illustrates a comparison between \dataset{} and existing datasets for step sequence generation/prediction tasks. \dataset{} is task-oriented and multimodal, encompassing both text and video data. The textual component involves the task goal and text descriptions for each step towards that goal. To support research on the level of task step, the visual content in \dataset{} includes video demonstrations of both the complete task and its individual steps.
\begin{table*}[th]
\resizebox{0.95\textwidth}{!}{%
\begin{tabular}{c|ccccccc}
\hline \hline
                          & \textbf{Data Type} & \textbf{Entire/Clip Video} & \textbf{Step Image} & \textbf{Preceding Steps} & \textbf{Subsequent Steps} & \textbf{Future Steps} & \textbf{Task Goal} \\ \hline
\textbf{RobotHow}         & Text               & \CheckmarkBold / \XSolidBrush                            & \XSolidBrush                   & \CheckmarkBold                         & \CheckmarkBold                   & \CheckmarkBold                     & \CheckmarkBold                  \\
\textbf{wikihow-GOSC}     & Text               & \XSolidBrush / \XSolidBrush                             & \XSolidBrush                   & \CheckmarkBold                         & \CheckmarkBold                   & \CheckmarkBold                     & \CheckmarkBold                  \\
\textbf{EpicKitchens-100} & Video              & \CheckmarkBold / \XSolidBrush                             & \CheckmarkBold                   & \XSolidBrush                         & \XSolidBrush                   & \XSolidBrush                     & \XSolidBrush                  \\
\textbf{EGTEA Gaze+}      & Video              & \CheckmarkBold / \CheckmarkBold                             & \CheckmarkBold                   & \XSolidBrush                         & \XSolidBrush                   & \XSolidBrush                     & \XSolidBrush                  \\
\textbf{wikiHow-VGSI}     & Multimedia         & \XSolidBrush / \XSolidBrush                             & \CheckmarkBold                   & \CheckmarkBold                         & \CheckmarkBold                   & \CheckmarkBold                     & \CheckmarkBold                  \\
\textbf{Coin}             & Multimedia         & \CheckmarkBold / \CheckmarkBold                             & \CheckmarkBold                   & \CheckmarkBold                         & \CheckmarkBold                   & \CheckmarkBold                     & \CheckmarkBold                  \\
\textbf{HowTo100M}        & Multimedia         & \CheckmarkBold / \CheckmarkBold                             & \CheckmarkBold                   & \CheckmarkBold                         & \CheckmarkBold                   & \CheckmarkBold                     & \CheckmarkBold                  \\
\textbf{\dataset{}}            & Multimedia         & \CheckmarkBold / \CheckmarkBold                             & \CheckmarkBold                   & \CheckmarkBold                         & \CheckmarkBold                   & \CheckmarkBold                     & \CheckmarkBold                  \\ \hline \hline
\end{tabular}
}
\caption{Comparisons of existing step sequence generation/prediction datasets and \dataset{}. 
}
\label{tab:exist_dataset}
\end{table*} 

\subsection{Implementation Details}
For both approaches, we use a pre-trained UniVL~\cite{UniVL} as the backbone encoder and decoder. UniVL consists of a text encoder which is built on BERT, a video encoder as well as a cross encoder to capture the cross-modality interactions while both the video encoder and cross-encoder are based on Transformer. UniVL also consists of a transformer-based decoder. We commenced our fine-tuning with the UniVL checkpoint pre-trained on the \textit{YoucookII} dataset. The Deberta-based NLI model is based on the pre-trained checkpoint of ``\textit{nli-deberta-v3-base}''. During training for both tasks, we employ the Adam optimizer with a learning rate of 3e-5, a batch size of 128, and a maximum of 100 epochs. We also use an early stopping mechanism, halting the training process if there is no improvement in average performance on the development set over five consecutive epochs.

\end{document}